\documentclass[10pt,twocolumn,letterpaper]{article}

\usepackage[pagenumbers]{cvpr} 

\usepackage{multirow}

\definecolor{cvprblue}{rgb}{0.21,0.49,0.74}
\usepackage[pagebackref,breaklinks,colorlinks,allcolors=cvprblue]{hyperref}

\def\paperID{DRAFT} 
\def\confName{CVPR}
\def\confYear{2027}

\title{ConstraintMatch: Compositional Image--Text Matching through Constrained Regional Evidence Selection}

\author{Yuhong Zhou$^{1}$ \quad Yubo Gao$^{2}$ \quad Xiangbo Shu$^{1}$\thanks{Corresponding author.}\\
$^{1}$Nanjing University of Science and Technology\\
$^{2}$The Hong Kong University of Science and Technology (Guangzhou)\\
{\tt\small zhouyh0315@gmail.com \quad yubogao1015@gmail.com \quad shuxb@njust.edu.cn}
}

\begin{document}
\maketitle
\begin{abstract}
Local image--text matching can find strong evidence for individual concepts yet favor captions with incorrect attribute--entity bindings. We introduce ConstraintMatch, a training-free inference method that turns entity-level regional consistency into an explicit matching cost. On the same frozen condition--region similarity matrix, it compares independent crop selection for each category and attribute with a shared crop choice within each entity. Their gap quantifies the score advantage obtained by relaxing the shared-region constraint. Spatial references from the same image calibrate this gap by keeping category evidence fixed while jointly shifting an entity's attribute maps. Per-entity clipping prevents cancellation across residuals, and the scaled mean cost is subtracted from global similarity. ConstraintMatch uses pooled embeddings, and spatial calibration requires no additional encoder calls. With parameters fixed per backbone across tasks, ConstraintMatch improves over global matching by 7.70 percentage points on full ARO Attribute and 5.63 points on held-out SugarCrepe++ attribute swaps, averaged across Qwen~2B and CORE~2B. Against SGI adapted to the same 13-crop budget, the respective gains are 4.85 and 3.51 points. Controlled analyses distinguish the effects of shared-region matching and candidate score calibration. The results establish entity-level evidence selection as a practical scoring principle for compositional multimodal retrieval.
\end{abstract}

\section{Introduction}
\label{sec:intro}

\begin{figure}[t]
    \centering
    \includegraphics[width=\linewidth]{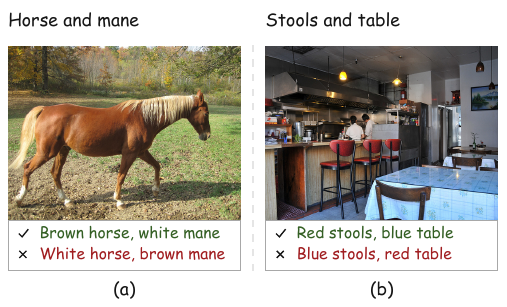}
    \caption{Attribute binding in SCPP~\cite{cm_scpp}. Checks/crosses mark positive/swapped captions with the same entities and colors. Text is shortened; (b) is cropped for display.}
    \label{fig:binding_examples}
\end{figure}

Recognizing the objects and attributes in a scene does not by itself establish whether a caption assigns them correctly. In \cref{fig:binding_examples}, \emph{a brown horse with a white mane} and \emph{a white horse with a brown mane} contain the same concepts but specify different bindings. The restaurant example makes the same distinction across objects: \emph{red stools and a blue table} versus \emph{blue stools and a red table}. Distinguishing these descriptions requires evidence for the relationship between an attribute and its entity, making attribute binding a central challenge in compositional image--text matching~\cite{cm_aro,cm_winoground,cm_sgi}.

Global embeddings summarize scene content but can overlook correspondences between individual conditions~\cite{cm_clip,cm_align,cm_aro,cm_crepe}. Recent approaches address this issue through phrase-level crop selection (SGI)~\cite{cm_sgi}, refinement of object and attribute representations (ABE-CLIP)~\cite{cm_abe}, and explicit constraint execution over concept evidence (LCSE)~\cite{cm_lcse}. These approaches motivate a closer look at evidence selection within an entity. We ask how much the matching score increases when category and attribute conditions may select regions independently instead of sharing one region, and whether this advantage can inform compositional matching.

The distinction matters because strong individual matches can support an incorrect composition. For the phrase \emph{blue stools} in \cref{fig:binding_examples}, category evidence could come from the stools while color evidence comes from the table. Rewarding these separate matches does not establish that the stools are blue. Comparing the two selection rules on identical evidence quantifies the advantage of allowing such separate choices, without attributing differences to a changed encoder or a different crop pool.

We introduce \textbf{ConstraintMatch}, a training-free method that turns this selection advantage into an entity-level correction to global similarity. Frozen encoders produce a condition--region similarity matrix from fixed image crops and separate category and attribute descriptions. Independent selection finds the best crop for each condition before averaging; shared selection averages condition evidence before choosing a crop. Sharing applies within an entity, while different entities retain separate choices. Their difference measures the score advantage obtained by relaxing the shared-region requirement, giving the correction an explicit interpretation in terms of evidence selection.

Fixed crops provide an approximation to object consistency: a crop can contain multiple objects, and an object's conditions can favor different crops. ConstraintMatch calibrates this variation using spatial references from the same image. Category score maps remain fixed while an entity's attribute maps shift together, preserving their mutual alignment. The reference adjusts the observed gap, and each entity's residual is clipped at zero before averaging, with zero-cost entities retained. The scaled mean cost is subtracted from global similarity (Figure~\ref{fig:cm-overview}). All operations use standard pooled embeddings; spatial calibration reuses existing similarities without further encoder calls.

With one parameter setting fixed per backbone, ConstraintMatch improves over global matching by 7.70 percentage points on full ARO Attribute and 5.63 points on held-out SugarCrepe++ (SCPP) attribute swaps, averaged across Qwen~2B and CORE~2B. It also outperforms SGI at the same regional budget on these binding tasks. COLA is used in development; we reevaluate the previously analysed ARO and held-out SCPP benchmarks with the method and parameters fixed. We additionally examine bidirectional matching, broader task families and inference cost to relate the gains to task structure and the evidence available to the method.

Our contributions are:
\begin{itemize}
    \item An entity-level correction that compares independent and shared selection on identical frozen evidence, calibrates their gap with same-image spatial references, and retains nonnegative entity costs without additional training.
    \item An analysis separating shared-region discrimination from candidate score calibration, and clarifying how the independent reference, spatial adjustment, entity aggregation and parsing support shape compositional matching.
    \item Attribute-binding gains from ConstraintMatch over global matching and budget-matched SGI on two backbones, with evaluations of bidirectional matching, transfer and computational tradeoffs.
\end{itemize}

\section{Related Work}
\label{sec:related}

\paragraph{Image--text retrieval and pretrained alignment.}
CLIP and ALIGN learn global image--text alignment for retrieval and zero-shot transfer~\cite{cm_clip,cm_align}. LiT aligns text with a locked image encoder~\cite{cm_lit}, while SigLIP uses a pairwise sigmoid objective~\cite{cm_siglip} and ALBEF and BLIP combine alignment with multimodal learning~\cite{cm_albef,cm_blip}. We study evidence selection and aggregation over fixed pretrained representations.

\paragraph{Compositional matching and attribute binding.}
Compositional evaluation separates concept recognition from structural matching. ARO probes attributes, relations and word order~\cite{cm_aro}, while VL-CheckList separates object, attribute and relation tests~\cite{cm_vlchecklist}. Caption-based tests address linguistic shortcuts through SugarCrepe's controlled negatives~\cite{cm_sugarcrepe} and SugarCrepe++'s semantic changes and equivalent lexical variations~\cite{cm_scpp}. Winoground and COLA test structure through contrasting image--caption combinations~\cite{cm_winoground,cm_cola}, and EqBen emphasizes minimal visual changes~\cite{cm_eqsim}. VALSE examines linguistic grounding~\cite{cm_valse}, CREPE tests systematicity and productivity~\cite{cm_crepe}, and related analyses examine composition within visual embeddings~\cite{cm_gde}.

Training strengthens composition through hard negatives in NegCLIP~\cite{cm_aro}, language-structured augmentation in SVLC~\cite{cm_svlc}, synthetic negative captions and images in TripletCLIP~\cite{cm_tripletclip}, scene graphs in Structure-CLIP~\cite{cm_structureclip}, and improved caption guidance~\cite{cm_semanticguidance}. Our intervention operates at inference time with fixed representations.

\paragraph{Local alignment and object-centric binding.}
Region--word retrieval models learn local alignments through cross-attention, regional reasoning and similarity filtering~\cite{cm_visualalign,cm_scan,cm_vsrn,cm_sgraf}. During pretraining, FILIP uses token-wise maximum patch similarity~\cite{cm_filip}, whereas FLAIR learns text-conditioned attention pooling from detailed descriptions and sub-captions~\cite{cm_flair}. Multigranular alignment in X-VLM~\cite{cm_xvlm} and the detailed captions, region supervision and fine-grained negatives in FG-CLIP~\cite{cm_fgclip} shape local evidence.

Object-centric methods model binding explicitly. OC-CLIP aligns caption scene graphs with visual slots through competitive attention and a learned structured similarity~\cite{cm_occlip}; CORA composes attributes and relations through scene-graph attention with global and local alignment losses~\cite{cm_cora}. In scene-memory retrieval, FRAME learns attribute readouts over stored object representations and aggregates them by query~\cite{cm_frame}. ConstraintMatch imposes an entity-level shared-selection constraint on fixed crop embeddings without learning a binding module. A shared crop approximates object consistency and may contain several objects. ConstraintMatch forms its score correction by comparing independent and shared region selection within each entity.

\paragraph{Structured inference and score correction.}
Inference methods differ in their evidence and decision rules. ComCLIP combines weighted subject/object/action subimage embeddings with the global representation without further training~\cite{cm_comclip}. Grounding CLIP uses an open-vocabulary detector to localize entities and relations before fusing local and global embeddings~\cite{cm_groundingclip}. SGI averages best-crop similarities for complete object--attribute phrases and the full caption~\cite{cm_sgi}; its paper separately trains an alignment transformer over frozen patch and token features. SGI preserves binding within each composed phrase when selecting a crop. ConstraintMatch instead encodes category and attribute conditions separately and ties their crop choices in the shared score. Comparing independent and shared choices on this matrix measures the score advantage obtained by allowing conditions to select regions independently.

Representation refinement and logical execution provide related routes to score correction. ABE-CLIP refines object and attribute text representations, uses token--patch alignment, and combines the refined local score and its absolute change from the original score with global similarity~\cite{cm_abe}. LCSE separates evidence extraction from constraint execution in a framework extending beyond Boolean constraints~\cite{cm_lcse}; its implementation corrects global similarity using the logit difference between logical aggregation of calibrated concept evidence and ordinary averaging. We compare selection constraints on identical condition--region evidence, calibrate their gap with spatial references, and analyse when the independent term cancels between candidates. We use SGI (13 regions) for matched-backbone quantitative comparisons, and ABE-CLIP and LCSE to contrast representation refinement and logical score correction.

\section{Method}
\label{sec:method}

\begin{figure*}[t]
\centering
\includegraphics[width=\textwidth]{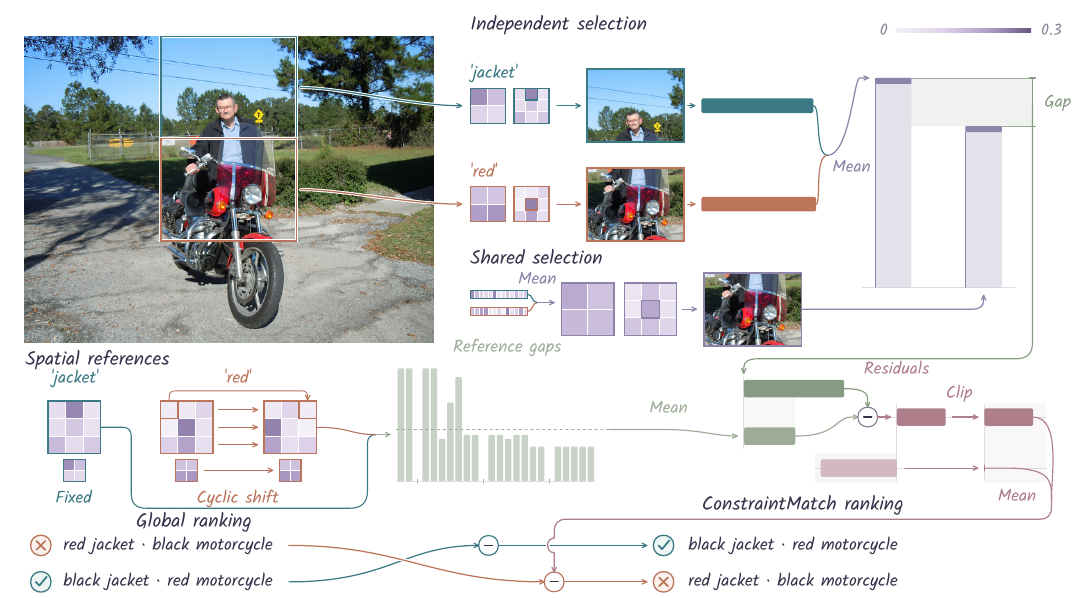}
\caption{\textbf{ConstraintMatch overview.} Frozen category and attribute evidence is aggregated under independent and shared selection. The expanded computation follows the swapped candidate's \emph{red jacket} entity ($U_e,C_e,\Delta_e,B_e$); the second residual row, \emph{black motorcycle}, uses its own gap and reference. Bold cells mark maxima in the selection panels and track a permuted score entry in the spatial-reference panels. Photo frames are actual fixed-grid crops; purple grids show cosine similarities. Residuals are clipped per entity before averaging, including zeros. In this Qwen~2B candidate pair, subtracting the swapped caption's cost reverses the ranking; the positive caption has zero cost. Bottom phrases are abbreviated, while Global uses the whole image and full effective captions. Checks/crosses indicate dataset positive/negative annotations.}
\label{fig:cm-overview}
\end{figure*}

ConstraintMatch compares independent and shared region selection on fixed evidence, calibrates their gap with spatial references from the same image, and averages nonnegative entity costs to correct global similarity (Figure~\ref{fig:cm-overview}). All encoders remain frozen.

\subsection{Representations and Entity Evidence}
\label{sec:method-evidence}

For an image $I$ and caption $T$, the frozen encoding interfaces $f_I$ and $f_T$ return $\ell_2$-normalized pooled vectors in $\mathbb{R}^d$. Their global similarity is $G(I,T)=f_I(I)^\top f_T(T)$, using the whole image and full effective caption.

Fixed dependency-parsing rules extract supported entities $\mathcal{E}(T)$, with $M_T=|\mathcal{E}(T)|$. Each entity $e$ contains an object category and its directly modifying attributes, represented by $K_e$ atomic descriptions $\mathcal{A}_e=\{a_{e1},\ldots,a_{eK_e}\}$. Category and attribute descriptions are encoded separately with fixed prompts and equal within-entity weights. This exposes their individual region preferences while preserving modifier relationships: \emph{black jacket} and \emph{red motorcycle} in Figure~\ref{fig:cm-overview} form distinct entities. The global branch retains caption content outside these groups. Appendix~A.1 specifies the prompts, input policy and parsing rules.

A caption-independent $2\times2$ grid and $3\times3$ grid provide $R=13$ crops $\mathcal{R}(I)=\{r_1,\ldots,r_R\}$. Keeping the whole image outside this local set reduces the chance that a view containing several objects masks disagreement among region choices. The condition--crop evidence for each entity is
\begin{equation}
A_e[k,j]=f_T(a_{ek})^\top f_I(r_j),
\qquad A_e\in\mathbb{R}^{K_e\times R}.
\label{eq:cm-evidence}
\end{equation}
Rows index conditions and columns index crops. All selection and calibration operations reuse this matrix and the same fixed representations.

\subsection{Shared-Region Gap Correction}
\label{sec:method-gap}

For a fixed pair $(I,T)$, we omit its arguments from entity-level scores. With region indices $j\in\{1,\ldots,R\}$, independent and shared selection are
\begin{equation}
\begin{aligned}
U_e&=\frac{1}{K_e}\sum_{k=1}^{K_e}\max_j A_e[k,j],\\
C_e&=\max_j\frac{1}{K_e}\sum_{k=1}^{K_e}A_e[k,j].
\end{aligned}
\label{eq:cm-selection}
\end{equation}
Independent selection chooses a region for each condition before averaging; shared selection averages condition evidence before choosing a region. Sharing applies within each entity: different entities may select different crops or reuse a crop.

Their gap measures the cost of requiring an entity's conditions to share regional evidence. For $M_T>0$, we correct global similarity by its entity average:
\begin{equation}
\begin{aligned}
\Delta_e&=U_e-C_e\ge0,\\
S_{\mathrm{base}}(I,T)&=G(I,T)-\frac{\lambda}{M_T}
\sum_{e\in\mathcal{E}(T)}\Delta_e .
\end{aligned}
\label{eq:cm-base}
\end{equation}
A small gap indicates that a common region nearly matches the separate optima; larger gaps record the advantage of independent choices. The correction rewards shared compatibility while calibrating it against the independently attainable score. Both use the same atomic similarities, isolating the selection constraint from the representation. Appendix~A gives the equivalent score decomposition. We use equal entity weights and $\lambda=2$. Captions with no supported entity, or triggering predefined caption-level fallbacks, retain $G(I,T)$.

\subsection{Spatial Reference and Per-Entity Adjustment}
\label{sec:method-spatial}

\paragraph{Spatial references.}
We calibrate each gap using spatial shifts of the same evidence. Within an $n\times n$ grid, a cyclic shift adds an offset $(\delta_y,\delta_x)\in\{0,\ldots,n-1\}^2$ to the region coordinates modulo $n$. Excluding the zero offset separately at each scale leaves three shifts for the $2\times2$ grid and eight for the $3\times3$ grid. Their Cartesian product defines a fixed set $\mathcal{T}$ of 24 transformations, each permuting indices only within its scale.

Let row 1 of $A_e$ contain category evidence and the remaining rows contain attributes. For the region permutation $\pi_\tau$ induced by $\tau\in\mathcal{T}$, define
\begin{equation}
A_e^{(\tau)}[k,j]=
\begin{cases}
A_e[1,j], & k=1,\\
A_e[k,\pi_\tau(j)], & k>1.
\end{cases}
\label{eq:cm-spatial-matrix}
\end{equation}
All attributes of an entity use the same transformation, preserving their mutual alignment and each row's score distribution while changing their correspondence with the fixed category evidence. Row maxima are preserved, so $U_e$ is unchanged and only shared compatibility must be recomputed. Shifting all conditions together would merely relabel regions and leave both scores unchanged; anchoring the category makes the reference sensitive to category--attribute correspondence.

\paragraph{Calibrated entity costs.}
We recompute shared scores and average their gaps from $U_e$:
\begin{equation}
\begin{aligned}
C_e^{(\tau)}&=\max_j\frac{1}{K_e}
\sum_{k=1}^{K_e}A_e^{(\tau)}[k,j],\\
B_e&=\frac{1}{|\mathcal{T}|}\sum_{\tau\in\mathcal{T}}
\bigl(U_e-C_e^{(\tau)}\bigr).
\end{aligned}
\label{eq:cm-spatial-reference}
\end{equation}
Each reference $B_e\ge0$ uses the same atoms, weights and similarities as its observed gap $\Delta_e$. Computing these references requires only matrix indexing and aggregation, with no additional image or atomic-text encoding.

We subtract a fraction $\eta\in[0,1]$ of the reference from each entity's gap, clip its residual, and average the nonnegative costs:
\begin{equation}
\begin{aligned}
\widehat\Delta_e&=[\Delta_e-\eta B_e]_+,\\
S(I,T)&=G(I,T)-\frac{\lambda}{M_T}
\sum_{e\in\mathcal{E}(T)}\widehat\Delta_e .
\end{aligned}
\label{eq:cm-final}
\end{equation}
Here $[z]_+=\max(z,0)$. Clipping each entity before averaging prevents negative residuals from offsetting positive costs. All $M_T$ supported entities remain in the denominator, including those assigned zero cost, preserving their equal weights. Clipping after averaging permits cancellation when residuals have mixed signs; the two orders coincide for a single entity.

At $\eta=0$, the score recovers $S_{\mathrm{base}}$, with $S_{\mathrm{base}}\le S\le G$. At $\eta=1$, the adjusted cost measures the shortfall of observed shared compatibility relative to its mean under spatial shifts. Calibration thus depends on the image and entity: equal observed gaps need not receive equal relief. Appendix~A derives these identities and the exact penalty reduction. Uniform-weakening and mean-matched controls test whether this dependence adds value beyond reducing average penalty strength. The same $\lambda=2$ and global fallback rules apply. We fix $\eta=1$ for Qwen~2B and $\eta=0.25$ for CORE~2B across tasks; Section~\ref{sec:exp-setup} specifies the development protocol.

The mixed-sign residuals in Figure~\ref{fig:cm-overview} illustrate the aggregation order: clipping retains the positive entity cost and sets the other cost to zero, with both entities included in the mean. Averaging signed residuals first would allow cancellation. For this pair, the resulting penalty reverses the swapped caption's initial ranking advantage.

\subsection{Inference and Computational Cost}
\label{sec:method-cost}

We cache one global and 13 regional vectors per image; identical atomic prompts share vectors while retaining entity memberships and weights. Text-to-image retrieval precomputes image vectors and parses and encodes the query caption and atoms, holding its groups fixed across images. Image-to-text retrieval precomputes caption and atomic vectors and groups, then constructs the query's global and regional image vectors once for all candidates, using each caption's groups. Both directions use the same score and retain the benchmark's candidate pools and tie rules. Fixed parsing and representations also permit reuse of $U_e,C_e,B_e$ across calibration settings.

Once condition--crop similarities are available, the 24 spatial references are computed by reindexing and aggregating these scores, without further image or atomic-text encoding. This adds scoring work but no additional image-vector storage. At equal dimension and precision, the global and regional image-vector cache occupies 14 times the global-only image-vector footprint. Table~\ref{tab:cm-cost} reports measured latency and storage for the base configuration; cached spatial-reference computation is reported separately in Appendix~F. Appendix~B provides the operation counts and full storage accounting.

\section{Experiments}
\label{sec:experiments}
We test whether constrained evidence selection improves attribute binding, distinguish the roles of its independent and spatial references, and examine the resulting behavior across task families and retrieval directions. All encoders remain frozen.

\begin{table*}[t]
\centering
\small
\setlength{\tabcolsep}{0pt}
\renewcommand{\arraystretch}{1.10}
\begin{tabular}{@{}c@{\hspace{4pt}}c@{\hspace{4pt}}c@{\hspace{2pt}}c@{\hspace{5pt}}c@{\hspace{2pt}}c@{\hspace{5pt}}c@{\hspace{2pt}}c@{\hspace{2pt}}c@{}}
\toprule
\makebox[0.08\textwidth][c]{\multirow{2}{*}[-3pt]{Backbone}} & \makebox[0.18\textwidth][c]{\multirow{2}{*}[-3pt]{Method}} & \multicolumn{2}{c}{ARO Attribute} & \multicolumn{2}{c}{SCPP held-out} & \multicolumn{3}{c}{COLA} \\[1pt]
\cmidrule(l{1pt}r{1pt}){3-4}\cmidrule(l{1pt}r{1pt}){5-6}\cmidrule(l{1pt}r{1pt}){7-9}
& & \makebox[\dimexpr0.105714\textwidth-3.714286pt\relax][c]{Full} & \makebox[\dimexpr0.105714\textwidth-3.714286pt\relax][c]{Macro} & \makebox[\dimexpr0.105714\textwidth-3.714286pt\relax][c]{Attr. swap} & \makebox[\dimexpr0.105714\textwidth-3.714286pt\relax][c]{All} & \makebox[\dimexpr0.105714\textwidth-3.714286pt\relax][c]{I2T} & \makebox[\dimexpr0.105714\textwidth-3.714286pt\relax][c]{T2I} & \makebox[\dimexpr0.105714\textwidth-3.714286pt\relax][c]{All four} \\[1pt]
\midrule
\addlinespace[1.5pt]
\multirow{5}{*}{Qwen~2B} & Global & 68.71 & 67.56 & 54.80 & 74.51 & \textbf{49.52} & 25.24 & 19.05 \\
 & SGI (13 regions) & 70.28 & 71.52 & 58.49 & 70.60 & 40.95 & \textbf{30.95} & \textbf{22.38} \\
 & Shared fusion & 71.07 & 71.04 & 59.04 & 68.71 & 32.38 & 24.76 & 13.33 \\
 & ConstraintMatch (base) & \textbf{75.89} & \textbf{76.80} & \textbf{62.92} & 75.01 & 39.05 & 28.10 & 15.71 \\
 & \textbf{ConstraintMatch} & 74.65 & 74.47 & 59.96 & \textbf{75.24} & 48.10 & 26.19 & 20.00 \\
\addlinespace[1.5pt]
\midrule
\addlinespace[1.5pt]
\multirow{5}{*}{CORE~2B} & Global & 67.21 & 66.72 & 65.68 & 79.73 & 48.57 & 32.38 & 23.81 \\
 & SGI (13 regions) & 71.34 & 73.21 & 66.24 & 75.96 & 47.62 & \textbf{40.95} & \textbf{29.52} \\
 & Shared fusion & 72.27 & 73.38 & 66.61 & 72.84 & 35.71 & 32.86 & 16.19 \\
 & ConstraintMatch (base) & 76.23 & 77.41 & 71.03 & 79.62 & 50.48 & 32.86 & 25.71 \\
 & \textbf{ConstraintMatch} & \textbf{76.67} & \textbf{78.16} & \textbf{71.77} & \textbf{80.07} & \textbf{50.95} & 35.24 & 27.62 \\
\bottomrule
\end{tabular}
\caption{\textbf{Comparison at a matched regional budget.} Accuracy (\%). Exact pointwise 95\% bootstrap intervals are given in Appendix~C.3. Regional methods use 13 crops; our controls and full method use $\lambda=2$. The full method fixes $\eta=1/0.25$ for Qwen~2B/CORE~2B. COLA is used in development. Bold denotes the best displayed accuracy per backbone.}
\label{tab:cm-main}
\end{table*}

\subsection{Experimental Setup}
\label{sec:exp-setup}
\paragraph{Benchmarks.}
Our main binding evaluations are full ARO Attribute~\cite{cm_aro} and the source-held-out partition of SugarCrepe++ (SCPP)~\cite{cm_scpp}. We report both full and official macro accuracy for ARO, and both attribute-swap and overall accuracy for SCPP. The mechanism analysis uses a separate source-filtered ARO population. COLA multi-object matching~\cite{cm_cola} evaluates two images and two captions: I2T and T2I require both matches within a direction, while all-four requires all four strict inequalities. Broader evaluations cover ARO Relation and order sensitivity, COLA single-object retrieval, and Flickr30k~\cite{cm_flickr}. Appendix~C gives the population sizes, splits and metric rules.

\paragraph{Backbones and controls.}
We use Qwen3-VL-Embedding-2B~\cite{cm_qwen} (Qwen~2B) and CORE-Embedding-2B~\cite{cm_core} (CORE~2B), with SigLIP~2 L/16-384~\cite{cm_siglip2} in additional mechanism and budget studies. Global uses whole-image and full-caption embeddings. Our matched-budget adaptation of SGI~\cite{cm_sgi} shares the backbone, 13 local crops and global view with ConstraintMatch. This fixes the available visual observations while retaining each method's text decomposition and aggregation; it does not equate their text-encoding or scoring costs. Shared fusion additionally fixes the condition--region similarities, providing a more focused control for the selection rule. Reported two-backbone means are arithmetic averages.

\paragraph{Selection and uncertainty.}
We fix one parameter setting per backbone across tasks. The coefficient $\lambda=2$ was selected on SCPP development swaps. Spatial calibration is selected from $\eta\in\{0,0.25,0.5,1\}$: eligible values must match or exceed base accuracy on both the full development partition and its attribute-swap subset, after which we maximize COLA I2T and break ties toward smaller $\eta$. This gives $\eta=1$ for Qwen~2B and $\eta=0.25$ for CORE~2B. We then freeze these settings across tasks and reevaluate the previously analysed ARO and held-out SCPP benchmarks; COLA remains development evidence. We report pointwise 95\% source-group bootstrap intervals from 2,000 resamples. Paired comparisons use the same resampled groups for both methods. Evaluation is deterministic, so these intervals measure resampling uncertainty rather than training-seed variation. Appendix~C.2 provides the complete protocol and paired results.

\begin{figure}[t]
\centering
\includegraphics[width=\linewidth]{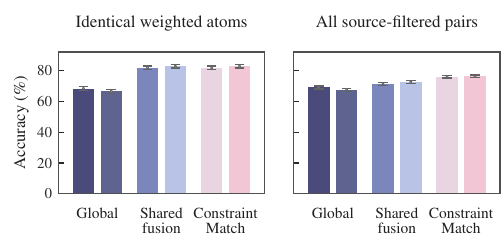}
\caption{\textbf{Shared selection and independent-reference calibration.} Accuracy on source-filtered ARO for the base configuration ($\eta=0$, $\lambda=2$, 13 crops). Left: 9,295 pairs with identical weighted atoms; right: all 26,598 filtered pairs. Within each method, bars show Qwen~2B then CORE~2B. Whiskers show pointwise 95\% source-group percentile bootstrap intervals (2,000 resamples). Methods are compared within each population and backbone; the left panel is a subset of the right.}
\label{fig:cm-structure}
\end{figure}

\begin{table}[t]
\centering
\small
\setlength{\tabcolsep}{5pt}
\renewcommand{\arraystretch}{1.10}
\begin{tabular}{@{}c@{}c*{3}{@{}c}@{}}
\toprule
\makebox[0.17\linewidth][c]{Backbone} & \makebox[0.37\linewidth][c]{Variant} & \makebox[0.15\linewidth][c]{ARO} & \makebox[0.15\linewidth][c]{SCPP} & \makebox[0.16\linewidth][c]{COLA} \\
\midrule
\addlinespace[1pt]
\multirow{6}{*}{Qwen~2B} & w/o spatial ref. & 75.89 & 62.92 & 15.71 \\
 & Uniform weakening & 68.71 & 54.80 & 19.05 \\
 & Mean-matched & 73.77 & 56.64 & 18.57 \\
 & w/o clipping & \textbf{77.00} & \textbf{63.47} & 18.57 \\
 & Post-mean clip & 74.07 & 59.59 & \textbf{20.00} \\
 & \textbf{ConstraintMatch} & 74.65 & 59.96 & \textbf{20.00} \\
\addlinespace[1pt]
\midrule
\addlinespace[1pt]
\multirow{6}{*}{CORE~2B} & w/o spatial ref. & 76.23 & 71.03 & 25.71 \\
 & Uniform weakening & 76.17 & 70.85 & 25.71 \\
 & Mean-matched & 76.17 & 70.85 & 25.71 \\
 & w/o clipping & \textbf{77.21} & 71.96 & \textbf{27.62} \\
 & Post-mean clip & 76.87 & \textbf{72.14} & 27.14 \\
 & \textbf{ConstraintMatch} & 76.67 & 71.77 & \textbf{27.62} \\
\bottomrule
\end{tabular}
\caption{\textbf{Component ablations.} Accuracy (\%). ARO: full ARO Attribute; SCPP: held-out attribute swaps; COLA: all-four matching. All variants use 13 crops and $\lambda=2$; spatial variants use the fixed $\eta$ from Table~\ref{tab:cm-main}. Without spatial references, $\eta=0$. Uniform weakening and Mean-matched rescale the original gap. Removing clipping retains signed entity residuals; Post-mean clip clips after entity averaging, allowing cancellation; ConstraintMatch clips each entity before averaging. COLA is used in development. Bold denotes the best displayed value per backbone and metric; paired comparisons appear in Appendix~C.4.}
\label{tab:cm-ablation}
\end{table}

\subsection{Attribute Binding and Bidirectional Matching}
\label{sec:exp-main}
ConstraintMatch improves attribute binding on both backbones relative to Global, shared fusion and matched-budget SGI (Table~\ref{tab:cm-main}). Averaged across Qwen~2B and CORE~2B, ConstraintMatch gains 7.70 percentage points on full ARO Attribute and 5.63 points on held-out SCPP attribute swaps over Global. Improvements also hold under ARO's official macro-average, which uses its eligible attribute-pair subset and weights groups equally. The same scoring rule thus retains gains under both aggregate and group-balanced evaluation. Its advantage over SGI at the same crop budget further indicates that access to regional views alone does not account for the improvement: how their evidence is selected and compared matters.

These comparisons probe different aspects of the score. Global tests whether the correction helps beyond whole-scene similarity; matched-budget SGI tests whether the improvement is explained by adding the same regional views; the base method versus shared fusion tests the independent reference while fixing atomic similarities. This distinction matters for attribute swaps, where both candidates can match the image's inventory of concepts. Recognizing each concept alone does not establish which entity it modifies: the score must respond to how the conditions are grouped.

Bidirectional matching reveals a different balance between configurations. Spatial calibration raises COLA all-four accuracy over the base method and Global on both backbones, while SGI remains stronger on this criterion. Relative to Global, CORE~2B improves in both retrieval directions, whereas Qwen~2B remains slightly lower on I2T. Requiring all four matches tests the joint ranking of both images and captions, rather than only a single positive--negative caption decision. The joint-ranking improvement comes with lower attribute accuracy than the base configuration on Qwen~2B, whereas CORE~2B improves on both binding tasks. These are development observations, and the paired all-four gains over Global have intervals that include zero. We therefore use COLA to characterize the observed tradeoff, with the fixed-configuration ARO and SCPP reevaluations providing the evidence for attribute-binding gains.

\subsection{Mechanism and Ablations}
\label{sec:exp-ablation}
\paragraph{Shared selection and its reference.}
Shared compatibility and the independent reference have distinguishable roles. Figure~\ref{fig:cm-structure} compares shared fusion $G+2C$ with the base score $G+2C-2U$, where $C$ and $U$ are entity averages. When two captions have identical atomic prompts and cumulative weights, $U$ is identical and cancels from their ranking. Shared fusion and the base method then make exactly the same decisions across all three encoders, and both improve over Global. Attribute reassignment can change which conditions must share a region even when this independent evidence is unchanged.

The independent reference contributes when weighted atomic compositions differ. Across the full source-filtered population, subtracting $2U$ improves over shared fusion by 4.67 percentage points on Qwen~2B and 3.81 points on CORE~2B. The advantage persists in the subset where both captions have parsing support but different weighted atoms. It is therefore not solely explained by one candidate receiving a correction while the other falls back to Global. Together, the two populations distinguish compatibility under a common-region constraint from calibration against the score available under independent choices. All comparisons reuse the same fixed-crop evidence; complete estimates and paired intervals appear in Appendix~C.4.

\paragraph{Spatial calibration.}
The spatial controls test whether the reference helps only by weakening the penalty (Table~\ref{tab:cm-ablation}). Uniform weakening rescales every gap by the same factor. Mean-matched weakening instead matches the average adjusted penalty on unlabeled development inputs, while discarding its image- and entity-specific allocation. It therefore controls for average penalty strength on those inputs, allowing the comparison to examine where penalty relief is applied. Under either scalar control, equal original gaps receive equal adjusted costs, even when the underlying spatial alignments differ. The spatial reference can distinguish these cases by comparing each entity's observed alignment with its own shifted alignments. Mean matching fixes a common multiplier from development-set averages, so it cannot retain this entity-specific reference. This comparison therefore examines the allocation of penalty relief after controlling its average strength on development inputs.

After mean matching, spatial calibration improves ARO accuracy on both backbones. The paired analysis also supports the SCPP gain on Qwen~2B; the CORE~2B SCPP and both COLA intervals include zero. The ARO results thus support image- and entity-dependent calibration beyond matching the development-set mean penalty. These controls reuse the same atomic similarities and crops, so their differences do not rely on a stronger visual representation or additional regional evidence. The references preserve the independent optima while changing shared compatibility. Appendix~D gives the control definitions, fixed matching coefficients and complete paired comparisons.

\paragraph{Entity-level clipping.}
Clipping determines whether calibration acts only as a nonnegative inconsistency cost or can also reward an entity. Retaining signed residuals yields stronger attribute accuracy in Table~\ref{tab:cm-ablation}. ConstraintMatch instead preserves the cost interpretation: an entity with a negative residual contributes zero and cannot offset another entity's positive cost. Clipping after averaging permits this cancellation; clipping each entity first prevents it while retaining zero-cost entities in the denominator.

The accuracy effect of this ordering depends on the backbone and task. Compared with Post-mean clip, entity-level clipping improves Qwen~2B ARO accuracy but reduces CORE~2B ARO accuracy, and the two orders have the same mean SCPP swap accuracy. Their COLA comparison does not establish a positive gain. Together, these variants separate spatial calibration from the choice to aggregate nonnegative entity costs. The latter prevents cross-entity cancellation, while the measured tradeoffs identify how that constraint affects each task. Appendix~C.4 reports the corresponding intervals.

The base correction's binding gains also persist across the tested crop budgets and nonzero penalty coefficients on all three encoders. This sensitivity analysis tests whether the result depends on a single regional pool or coefficient choice. Appendix~D gives the complete sensitivity results and paired intervals.

\subsection{Task Structure and Transfer}
\label{sec:exp-transfer}
ConstraintMatch improves overall held-out SCPP accuracy on both backbones, with the largest gain concentrated in attribute swaps. Other SCPP task families change by less than one percentage point in the two-backbone mean. The overall score retains the benchmark's broader mixture of replacements and swaps, showing how that targeted improvement contributes to aggregate performance.

The difference between replacement and swapping makes this concentration interpretable. A replacement can change whether a concept is present, whereas an attribute swap preserves the concepts and challenges their assignment. Strong atomic matches can therefore support both the original caption and its attribute-swapped counterpart. Entity grouping allows shared-region compatibility to differ even when the independent score remains unchanged. This connects the task's construction to the selection constraint, explaining why attribute-swap accuracy is informative alongside the broader benchmark score.

We separately evaluate the base configuration on broader task families (Appendix~E). ARO Relation predictions coincide with Global, primarily because both captions trigger the unchanged fallback. Word-order accuracy decreases on COCO-Order and Flickr-Order. The error transitions identify an important interaction with parsing support: a supported positive can receive a penalty while a negative retains its uncorrected global score. This asymmetry affects candidate comparability as well as coverage and helps explain why binding gains do not automatically transfer to order-sensitive negatives. Single-object COLA and Flickr retrieval show smaller losses. These broader-task results concern the base configuration; Table~\ref{tab:cm-main} evaluates ConstraintMatch with spatial calibration.

A common-protocol SigLIP~2 comparison further separates the selection rule from crop count. The base method exceeds SGI with larger view budgets on source-filtered ARO and SCPP swaps, while SGI remains stronger on COLA and official ARO macro accuracy (Appendix~E.1). These results connect the value of the correction to task structure and available local evidence, rather than suggesting that increasing the regional budget resolves all compositional errors.

\subsection{Computational Cost}
\label{sec:exp-cost}
\begin{table}[!t]
\centering
\small
\renewcommand{\arraystretch}{1.10}
\setlength{\tabcolsep}{2pt}
\begin{tabular*}{\linewidth}{@{\extracolsep{\fill}}ccccc@{}}
\toprule
Backbone & Method & I2T & T2I & Index \\
\midrule
\multirow{2}{*}{Qwen~2B} & Global & 58.89 & 26.87 & 7.81 \\
 & ConstraintMatch (base) & 192.43 & 44.70 & 109.38 \\
\midrule
\multirow{2}{*}{CORE~2B} & Global & 58.65 & 29.13 & 7.81 \\
 & ConstraintMatch (base) & 193.51 & 60.09 & 109.38 \\
\bottomrule
\end{tabular*}
\caption{\textbf{Measured inference cost.} Median (p50) request latency (ms) and FP32 image-vector index size (MiB) for 1,000 images. I2T ranks 5,000 captions; T2I ranks 1,000 images using the cache-reset-pass protocol in Appendix~F. ConstraintMatch uses $\eta=0$; spatial-reference updates are measured separately.}
\label{tab:cm-cost}
\end{table}

Regional evidence construction adds image encoding and index storage beyond global matching (Table~\ref{tab:cm-cost}). The method stores one global and 13 regional vectors per image. Spatial calibration then reuses their condition--region similarities: it changes the reductions over existing evidence and requires neither further encoder calls nor additional image-vector copies. This distinction separates the cost of constructing regional representations from the cost of applying the calibrated score.

We measure the base method's end-to-end request latency and visual-index storage on one A800 SXM4 80GB GPU. The two retrieval directions follow their respective cache protocols, so their measured overheads need not be identical. We report cached spatial-reference updates separately to distinguish their matrix-level cost from end-to-end request latency. Appendix~F specifies the timing boundaries, numerical precision and cache handling, and reports the cached updates and larger-view SGI comparisons.

The retrieval direction determines where regional evidence is constructed. In image-to-text retrieval, regional vectors belong to the query image; in text-to-image retrieval, they belong to database images and are precomputed in the visual index. Reversing the direction therefore changes the query-time work while leaving the pairwise scoring rule unchanged. Request latency measures that workflow under the stated cache protocol. Index size instead measures the visual representations retained across requests, so it captures a storage requirement that latency alone does not describe. The index measurement includes image embeddings and excludes model weights, text caches and temporary working memory.

\section{Conclusion}
\label{sec:conclusion}

ConstraintMatch makes shared regional support for an entity's conditions an explicit component of image--text scoring. It compares independent and shared region choices on fixed evidence, then calibrates their gap using spatial references from the same image. Controlled analyses distinguish the use of a shared region to compare attribute assignments from the independent-reference correction used to calibrate candidate scores. On Qwen~2B and CORE~2B, ConstraintMatch improves attribute binding over global matching and budget-matched SGI, with the same scoring rule applicable in both retrieval directions. These findings show how structured evidence selection can complement frozen representations in compositional image--text matching without retraining the encoders.

\clearpage
{
    \small
    \bibliographystyle{ieeenat_fullname}
    \bibliography{main}

@String(AAAI = {AAAI})

@article{cm_sgi,
  title = {{Revisiting Compositionality in Dual-Encoder Vision-Language Models: The Role of Inference}},
  author = {Miranda, Imanol and Salaberria, Ander and Agirre, Eneko and Azkune, Gorka},
  journal = {arXiv preprint arXiv:2604.11496},
  year = {2026}
}

@article{cm_abe,
  title = {{ABE-CLIP: Training-Free Attribute Binding Enhancement for Compositional Image-Text Matching}},
  author = {Zhang, Qi and Chen, Yuxu and Deng, Lei and Shen, Lili},
  journal = {arXiv preprint arXiv:2512.17178},
  year = {2025}
}

@article{cm_lcse,
  title = {{Similarity Is Not Logic: Factored Inference for Dual-Encoder Vision-Language Models}},
  author = {Alshehri, Sultan and Yang, Zhantao and Zhang, Han and Savvides, Marios},
  journal = {arXiv preprint arXiv:2607.23052},
  year = {2026}
}

@inproceedings{cm_clip,
  title = {{Learning Transferable Visual Models From Natural Language Supervision}},
  author = {Radford, Alec and Kim, Jong Wook and Hallacy, Chris and Ramesh, Aditya and Goh, Gabriel and Agarwal, Sandhini and Sastry, Girish and Askell, Amanda and Mishkin, Pamela and Clark, Jack and Krueger, Gretchen and Sutskever, Ilya},
  booktitle = {International Conference on Machine Learning},
  year = {2021}
}

@inproceedings{cm_aro,
  title = {{When and Why Vision-Language Models Behave like Bags-of-Words, and What to Do About It?}},
  author = {Yuksekgonul, Mert and Bianchi, Federico and Kalluri, Pratyusha and Jurafsky, Dan and Zou, James},
  booktitle = {International Conference on Learning Representations},
  year = {2023}
}

@inproceedings{cm_sugarcrepe,
  title = {{SugarCrepe: Fixing Hackable Benchmarks for Vision-Language Compositionality}},
  author = {Hsieh, Cheng-Yu and Zhang, Jieyu and Ma, Zixian and Kembhavi, Aniruddha and Krishna, Ranjay},
  booktitle = {Advances in Neural Information Processing Systems},
  year = {2023}
}

@article{cm_scpp,
  title = {{SUGARCREPE++ Dataset: Vision-Language Model Sensitivity to Semantic and Lexical Alterations}},
  author = {Dumpala, Sri Harsha and Jaiswal, Aman and Sastry, Chandramouli and Milios, Evangelos and Oore, Sageev and Sajjad, Hassan},
  journal = {arXiv preprint arXiv:2406.11171},
  year = {2024}
}

@inproceedings{cm_winoground,
  title = {{Winoground: Probing Vision and Language Models for Visio-Linguistic Compositionality}},
  author = {Thrush, Tristan and Jiang, Ryan and Bartolo, Max and Singh, Amanpreet and Williams, Adina and Kiela, Douwe and Ross, Candace},
  booktitle = {Proceedings of the IEEE/CVF Conference on Computer Vision and Pattern Recognition},
  year = {2022}
}

@inproceedings{cm_cola,
  title = {{COLA: A Benchmark for Compositional Text-to-image Retrieval}},
  author = {Ray, Arijit and Radenovic, Filip and Dubey, Abhimanyu and Plummer, Bryan A. and Krishna, Ranjay and Saenko, Kate},
  booktitle = {Advances in Neural Information Processing Systems},
  year = {2023}
}

@inproceedings{cm_svlc,
  title = {{Teaching Structured Vision \& Language Concepts to Vision \& Language Models}},
  author = {Doveh, Sivan and Arbelle, Assaf and Harary, Sivan and Schwartz, Eli and Herzig, Roei and Giryes, Raja and Feris, Rogerio and Panda, Rameswar and Ullman, Shimon and Karlinsky, Leonid},
  booktitle = {Proceedings of the IEEE/CVF Conference on Computer Vision and Pattern Recognition},
  year = {2023}
}

@inproceedings{cm_tripletclip,
  title = {{TripletCLIP: Improving Compositional Reasoning of CLIP via Synthetic Vision-Language Negatives}},
  author = {Patel, Maitreya and Kusumba, Abhiram and Cheng, Sheng and Kim, Changhoon and Gokhale, Tejas and Baral, Chitta and Yang, Yezhou},
  booktitle = {Advances in Neural Information Processing Systems},
  year = {2024}
}

@inproceedings{cm_filip,
  title = {{FILIP: Fine-grained Interactive Language-Image Pre-Training}},
  author = {Yao, Lewei and Huang, Runhui and Hou, Lu and Lu, Guansong and Niu, Minzhe and Xu, Hang and Liang, Xiaodan and Li, Zhenguo and Jiang, Xin and Xu, Chunjing},
  booktitle = {International Conference on Learning Representations},
  year = {2022}
}

@inproceedings{cm_flair,
  title = {{FLAIR: VLM with Fine-grained Language-informed Image Representations}},
  author = {Xiao, Rui and Kim, Sanghwan and Georgescu, Mariana-Iuliana and Akata, Zeynep and Alaniz, Stephan},
  booktitle = {Proceedings of the IEEE/CVF Conference on Computer Vision and Pattern Recognition},
  year = {2025}
}

@article{cm_occlip,
  title = {{Object-centric Binding in Contrastive Language-Image Pretraining}},
  author = {Assouel, Rim and Bordes, Florian and Astolfi, Pietro and Drozdzal, Michal and Romero-Soriano, Adriana},
  journal = {arXiv preprint arXiv:2502.14113},
  year = {2025}
}

@article{cm_frame,
  title = {{FRAME: Factored Retrieval via Attribute Readouts for Object-Centric Scene Memory}},
  author = {Jeon, Woosang and Choi, Sanghyeok and Kim, Minwoo and Jung, Taehyun and Kim, Taehyeong},
  journal = {arXiv preprint arXiv:2609.08886},
  year = {2026}
}

@inproceedings{cm_comclip,
  title = {{ComCLIP: Training-Free Compositional Image and Text Matching}},
  author = {Jiang, Kenan and He, Xuehai and Xu, Ruize and Wang, Xin},
  booktitle = {Proceedings of the Conference of the North American Chapter of the Association for Computational Linguistics: Human Language Technologies},
  year = {2024}
}

@article{cm_qwen,
 title={{Qwen3-VL-Embedding and Qwen3-VL-Reranker: A Unified Framework for State-of-the-Art Multimodal Retrieval and Ranking}},
 author={Li, Mingxin and Zhang, Yanzhao and Long, Dingkun and Chen, Keqin and Song, Sibo and Bai, Shuai and Yang, Zhibo and Xie, Pengjun and Yang, An and Liu, Dayiheng and Zhou, Jingren and Lin, Junyang},
 journal={arXiv preprint arXiv:2601.04720}, year={2026}
}

@article{cm_core,
 title={{CORE: Improving Compositional Reasoning in MLLM Embedding via Reranker Distillation}},
 author={Song, Tingyu and Li, Mingxin and Zhang, Yanzhao and Long, Dingkun and Liu, Chu and Xie, Pengjun and Zhao, Yilun and Wu, Shu},
 journal={arXiv preprint arXiv:2609.04083}, year={2026}
}

@article{cm_siglip2,
 title={{SigLIP 2: Multilingual Vision-Language Encoders with Improved Semantic Understanding, Localization, and Dense Features}},
 author={Tschannen, Michael and Gritsenko, Alexey and Wang, Xiao and Naeem, Muhammad Ferjad and Alabdulmohsin, Ibrahim and Parthasarathy, Nikhil and Evans, Talfan and Beyer, Lucas and Xia, Ye and Mustafa, Basil and H{\'e}naff, Olivier and Harmsen, Jeremiah and Steiner, Andreas and Zhai, Xiaohua},
 journal={arXiv preprint arXiv:2502.14786}, year={2025}
}

@article{cm_flickr,
 title={From image descriptions to visual denotations: New similarity metrics for semantic inference over event descriptions},
 author={Young, Peter and Lai, Alice and Hodosh, Micah and Hockenmaier, Julia},
 journal={Transactions of the Association for Computational Linguistics}, volume={2}, pages={67--78}, year={2014}
}

@inproceedings{cm_align,
  title = {{Scaling Up Visual and Vision-Language Representation Learning With Noisy Text Supervision}},
  author = {Jia, Chao and Yang, Yinfei and Xia, Ye and Chen, Yi-Ting and Parekh, Zarana and Pham, Hieu and Le, Quoc and Sung, Yun-Hsuan and Li, Zhen and Duerig, Tom},
  booktitle = {International Conference on Machine Learning},
  year = {2021},
  pages = {4904--4916},
}

@inproceedings{cm_lit,
  title = {{LiT: Zero-Shot Transfer With Locked-Image Text Tuning}},
  author = {Zhai, Xiaohua and Wang, Xiao and Mustafa, Basil and Steiner, Andreas and Keysers, Daniel and Kolesnikov, Alexander and Beyer, Lucas},
  booktitle = {Proceedings of the IEEE/CVF Conference on Computer Vision and Pattern Recognition},
  year = {2022},
  pages = {18123--18133},
}

@inproceedings{cm_siglip,
  title = {{Sigmoid Loss for Language Image Pre-Training}},
  author = {Zhai, Xiaohua and Mustafa, Basil and Kolesnikov, Alexander and Beyer, Lucas},
  booktitle = {Proceedings of the IEEE/CVF International Conference on Computer Vision},
  year = {2023},
  pages = {11975--11986},
}

@inproceedings{cm_albef,
  title = {{Align before Fuse: Vision and Language Representation Learning with Momentum Distillation}},
  author = {Li, Junnan and Selvaraju, Ramprasaath and Gotmare, Akhilesh and Joty, Shafiq and Xiong, Caiming and Hoi, Steven Chu Hong},
  booktitle = {Advances in Neural Information Processing Systems},
  year = {2021},
}

@inproceedings{cm_blip,
  title = {{BLIP: Bootstrapping Language-Image Pre-training for Unified Vision-Language Understanding and Generation}},
  author = {Li, Junnan and Li, Dongxu and Xiong, Caiming and Hoi, Steven},
  booktitle = {International Conference on Machine Learning},
  year = {2022},
  pages = {12888--12900},
}

@inproceedings{cm_vlchecklist,
  title = {{An Explainable Toolbox for Evaluating Pre-trained Vision-Language Models}},
  author = {Zhao, Tiancheng and Zhang, Tianqi and Zhu, Mingwei and Shen, Haozhan and Lee, Kyusong and Lu, Xiaopeng and Yin, Jianwei},
  booktitle = {Proceedings of the 2022 Conference on Empirical Methods in Natural Language Processing: System Demonstrations},
  year = {2022},
  pages = {30--37},
}

@inproceedings{cm_valse,
  title = {{VALSE: A Task-Independent Benchmark for Vision and Language Models Centered on Linguistic Phenomena}},
  author = {Parcalabescu, Letitia and Cafagna, Michele and Muradjan, Lilitta and Frank, Anette and Calixto, Iacer and Gatt, Albert},
  booktitle = {Proceedings of the 60th Annual Meeting of the Association for Computational Linguistics (Volume 1: Long Papers)},
  year = {2022},
  pages = {8253--8280},
}

@inproceedings{cm_crepe,
  title = {{CREPE: Can Vision-Language Foundation Models Reason Compositionally?}},
  author = {Ma, Zixian and Hong, Jerry and Gul, Mustafa Omer and Gandhi, Mona and Gao, Irena and Krishna, Ranjay},
  booktitle = {Proceedings of the IEEE/CVF Conference on Computer Vision and Pattern Recognition},
  year = {2023},
  pages = {10910--10921},
}

@inproceedings{cm_eqsim,
  title = {{Equivariant Similarity for Vision-Language Foundation Models}},
  author = {Wang, Tan and Lin, Kevin and Li, Linjie and Lin, Chung-Ching and Yang, Zhengyuan and Zhang, Hanwang and Liu, Zicheng and Wang, Lijuan},
  booktitle = {Proceedings of the IEEE/CVF International Conference on Computer Vision},
  year = {2023},
  pages = {11998--12008},
}

@inproceedings{cm_structureclip,
  title = {{Structure-CLIP: Towards Scene Graph Knowledge to Enhance Multi-Modal Structured Representations}},
  author = {Huang, Yufeng and Tang, Jiji and Chen, Zhuo and Zhang, Rongsheng and Zhang, Xinfeng and Chen, Weijie and Zhao, Zeng and Zhao, Zhou and Lv, Tangjie and Hu, Zhipeng and Zhang, Wen},
  booktitle = {Proceedings of the AAAI Conference on Artificial Intelligence},
  year = {2024},
  pages = {2417--2425},
}

@inproceedings{cm_semanticguidance,
  title = {{Learning Visual Composition through Improved Semantic Guidance}},
  author = {Stone, Austin and Soltau, Hagen and Geirhos, Robert and Yi, Xi and Xia, Ye and Cao, Bingyi and Chen, Kaifeng and Ogale, Abhijit and Shlens, Jonathon},
  booktitle = {Proceedings of the IEEE/CVF Conference on Computer Vision and Pattern Recognition},
  year = {2025},
  pages = {3740--3750},
}

@inproceedings{cm_gde,
  title = {{Not Only Text: Exploring Compositionality of Visual Representations in Vision-Language Models}},
  author = {Berasi, Davide and Farina, Matteo and Mancini, Massimiliano and Ricci, Elisa and Strisciuglio, Nicola},
  booktitle = {Proceedings of the IEEE/CVF Conference on Computer Vision and Pattern Recognition},
  year = {2025},
  pages = {24917--24927},
}

@inproceedings{cm_visualalign,
  title = {{Deep Visual-Semantic Alignments for Generating Image Descriptions}},
  author = {Karpathy, Andrej and Fei-Fei, Li},
  booktitle = {Proceedings of the IEEE Conference on Computer Vision and Pattern Recognition},
  year = {2015},
  pages = {3128--3137},
}

@inproceedings{cm_scan,
  title = {{Stacked Cross Attention for Image-Text Matching}},
  author = {Lee, Kuang-Huei and Chen, Xi and Hua, Gang and Hu, Houdong and He, Xiaodong},
  booktitle = {Proceedings of the European Conference on Computer Vision},
  year = {2018},
}

@inproceedings{cm_vsrn,
  title = {{Visual Semantic Reasoning for Image-Text Matching}},
  author = {Li, Kunpeng and Zhang, Yulun and Li, Kai and Li, Yuanyuan and Fu, Yun},
  booktitle = {Proceedings of the IEEE/CVF International Conference on Computer Vision},
  year = {2019},
  pages = {4654--4662},
}

@inproceedings{cm_sgraf,
  title = {{Similarity Reasoning and Filtration for Image-Text Matching}},
  author = {Diao, Haiwen and Zhang, Ying and Ma, Lin and Lu, Huchuan},
  booktitle = {Proceedings of the AAAI Conference on Artificial Intelligence},
  year = {2021},
  pages = {1218--1226},
}

@inproceedings{cm_xvlm,
  title = {{Multi-Grained Vision Language Pre-Training: Aligning Texts with Visual Concepts}},
  author = {Zeng, Yan and Zhang, Xinsong and Li, Hang},
  booktitle = {International Conference on Machine Learning},
  year = {2022},
  pages = {25994--26009},
}

@inproceedings{cm_fgclip,
  title = {{FG-CLIP: Fine-Grained Visual and Textual Alignment}},
  author = {Xie, Chunyu and Wang, Bin and Kong, Fanjing and Li, Jincheng and Liang, Dawei and Zhang, Gengshen and Leng, Dawei and Yin, Yuhui},
  booktitle = {International Conference on Machine Learning},
  year = {2025},
  pages = {68777--68793},
}

@inproceedings{cm_cora,
  title = {{Composing Object Relations and Attributes for Image-Text Matching}},
  author = {Pham, Khoi and Huynh, Chuong and Lim, Ser-Nam and Shrivastava, Abhinav},
  booktitle = {Proceedings of the IEEE/CVF Conference on Computer Vision and Pattern Recognition},
  year = {2024},
  pages = {14354--14363},
}

@inproceedings{cm_groundingclip,
  title = {{Compositional Image-Text Matching and Retrieval by Grounding Entities}},
  author = {Vongala, Madhukar Reddy and Srivastava, Saurabh and Kosecka, Jana},
  booktitle = {Proceedings of the IEEE/CVF Conference on Computer Vision and Pattern Recognition Workshops},
  year = {2025},
  pages = {241--250},
}
}


\end{document}